\documentclass[%
  onecolumn
   , hidempi
]{mpi2015-cscpreprint}

\usepackage[american]{babel}

\usepackage{graphicx}

\usepackage{amssymb}
\usepackage{amsthm}

\usepackage{mymacros}
\usepackage{import}
\usepackage{xifthen}
\usepackage{pdfpages}
\usepackage{transparent}
\usepackage{cleveref}

\newcommand{\lqresnet}{{\sffamily{LQResNet}}}
\newcommand{\resnet}{{\sffamily{ResNet}}}
\newcommand{\resnets}{{\sffamily{ResNets}}}
\newcommand{\opinf}{{\sffamily{OpInf}}}
\begin{document}
  

\title{LQResNet: A Deep Neural Network Architecture for Learning Dynamic Processes}
  
\author[$\ast$]{Pawan Goyal}
\affil[$\ast$]{Max Planck Institute for Dynamics of Complex Technical Systems, 39106 Magdeburg, Germany.\authorcr
  \email{goyalp@mpi-magdeburg.mpg.de},
}
  
\author[$\dagger$]{Peter Benner}
\affil[$\dagger$]{Max Planck Institute for Dynamics of Complex Technical Systems, 39106 Magdeburg, Germany.\authorcr
  \email{benner@mpi-magdeburg.mpg.de}, 
}
  
\shorttitle{A Deep Neural Network Architecture for Learning Dynamic Processes}
\shortauthor{P. Goyal, P. Benner}
\shortdate{}
  

  
\abstract{%
Mathematical modeling is an essential step, for example, to analyze the transient behavior of a dynamical process and to perform engineering studies such as optimization and control.  With the help of first-principles and expert knowledge, a dynamic model can be built. However, for complex dynamic processes, appearing, e.g., in biology, chemical plants, neuroscience, financial markets, this often remains an onerous task.  Hence, data-driven modeling of the dynamics process becomes an attractive choice and is supported by the rapid advancement in sensor and measurement technology. 
A data-driven approach, namely operator inference framework, models a dynamic process, where a physics-informed structure of the nonlinear term is assumed.
In this work, we suggest combining the operator inference with certain deep neural network approaches to infer the unknown nonlinear dynamics of the system.
The approach uses recent advancements in deep learning and prior knowledge of the process if possible.  We briefly also discuss several extensions and advantages of the proposed methodology.
We demonstrate that the proposed methodology accomplishes the desired tasks for dynamics processes encountered in neural dynamics and the glycolytic oscillator. 
}
  
\maketitle

\section*{Introduction}
	With the rapid development in sensor and measurement technology, time-series data of processes have become available in large amounts with high accuracy. Machine learning and data science play an important role in analyzing and perceiving information of the underlying process dynamics from these data.  Building a model describing the dynamics is vital in designing and optimizing various processes, as well as predicting their long-term transient behavior. 
Inferring a dynamic process model from data, often called system identification, has a rich history; see, e.g., \cite{lennart1999system,VanM96}. While linear system identification is well established, nonlinear system identification is still far from being as good understood as for linear systems, despite having a similarly long research history, see, e.g., \cite{NarP90,SuyVdM96}. Nonlinear system identification often relies on a good hypothesis of the model; thus, it is not entirely a black-box technology. Fortunately, there are several scenarios where one can hypothesize a model structure based on a good understanding of the underlying dynamic behavior using expert knowledge or experience. 
Towards nonlinear system identification, a promising approach based on a symbolic regression was proposed \cite{bongard2007automated} to determine the potential structure of a nonlinear system. The method discovers a dynamical system solely from data. However, the approach is computationally expensive; thus, it is not suitable for a large-scale dynamical system.  Also, one needs to take care of the overfitting by using the Pareto front that balances model complexity and data fitting \cite{bongard2007automated}. 
If nonlinearities are only known to belong to a certain class of mathematical functions (a \emph{dictionary} of functions), sparse regression (see, e.g., \cite{tibshirani1996regression}) and compressive sensing (see, e.g., \cite{donoho2006compressed,candes2006robust}) based approaches have emerged as a kind of dictionary learning methods for nonlinear models, including dynamical systems. Here, it is observed that given a high-dimensional nonlinear function space, the process dynamics can often be accurately described by only a few terms from the dictionary \cite{wang2011predicting,schaeffer2013sparse, rudy2017data, ozolicnvs2013compressed, proctor2014exploiting, brunton2016discovering}.  Consequently, one obtains an interpretable and parsimonious process model. However, the success of these approaches highly depends on the quality of the constructed dictionary, i.e., the candidate nonlinear function space \cite{brunton2016discovering}. Thus, dictionaries need to be generated carefully using expert knowledge. In addition to these, there exist other techniques to infer a model of a dynamic process, for instance from time-series data \cite{crutchfield1987equations,kantz2004nonlinear},  equation-free modeling \cite{kevrekidis2003equation, ye2015equation}, dynamical models inference \cite{schmidt2011automated,daniels2015automated,daniels2015efficient}. In the category of equation-free modeling, dynamical mode decomposition \cite{rowley2009spectral,schmid2010dynamic, williams2015data} has also shown promising results. 
Moreover, deep learning (DL) approaches have been developed successfully for this task. They have the potential to build a model using only the time-history of dependent variables. For this, a deep neural network (DNN) is typically used, which can be understood as a composition of functions since the output of a layer is the input of the next layer. DNNs have shown phenomenal performance across various disciplines \cite{krizhevsky2012imagenet, lecun2015deep, goodfellow2016deep} (e.g., image classification, speech recognition, medical image analysis, to name a few). A primary reason for DNNs' success is that each layer learns a particular feature representation of the input, thus describing a complex function representation. Among several existing DNN architectures, \emph{Residual Neural Networks} (\resnet) \cite{he2016deep} closely resemble integration schemes (e.g., an Euler-type integration) for ordinary differential equations (ODEs). Inspired by this, there are methodological advances for inferring ODEs from data \cite{chen2018neural}. 

Furthermore, in recent years, scientific machine learning has emerged as a discipline that combines classical mechanistic modeling and numerical simulation with machine learning techniques. The expectation here is that incorporating existing scientific knowledge allows the (physical) interpretability of models while requiring less training data. Here, it is important to note that engineering processes often do not yield the same amount of data as typical socio-economic big data applications, so overfitting is often encountered. This can be attenuated by including physical constraints during learning, thereby reducing the network parameters or providing a better-suited network design. Operator inference approaches fall into this category; see, e.g., \cite{peherstorfer2016data, morBenGKPW20}. Furthermore, physics-informed neural networks (PINN) \cite{raissi2019physics} also employ the existing physical knowledge to improve the learning process that, as an output, yields the solution of a partial differential equation. However, as of yet, PINNs assume the structure of the physical model to be fully known, which is not available in a complex dynamic process. 

In the following section, we propose a DNN architecture dedicated to learning dynamical systems from data while incorporating certain structural assumptions implied by the underlying physics.

\section*{LQResNet: A deep  network  architecture for learning nonlinear dynamics}
\begin{figure*}[tb]
	\centering
	\includegraphics[width=\textwidth]{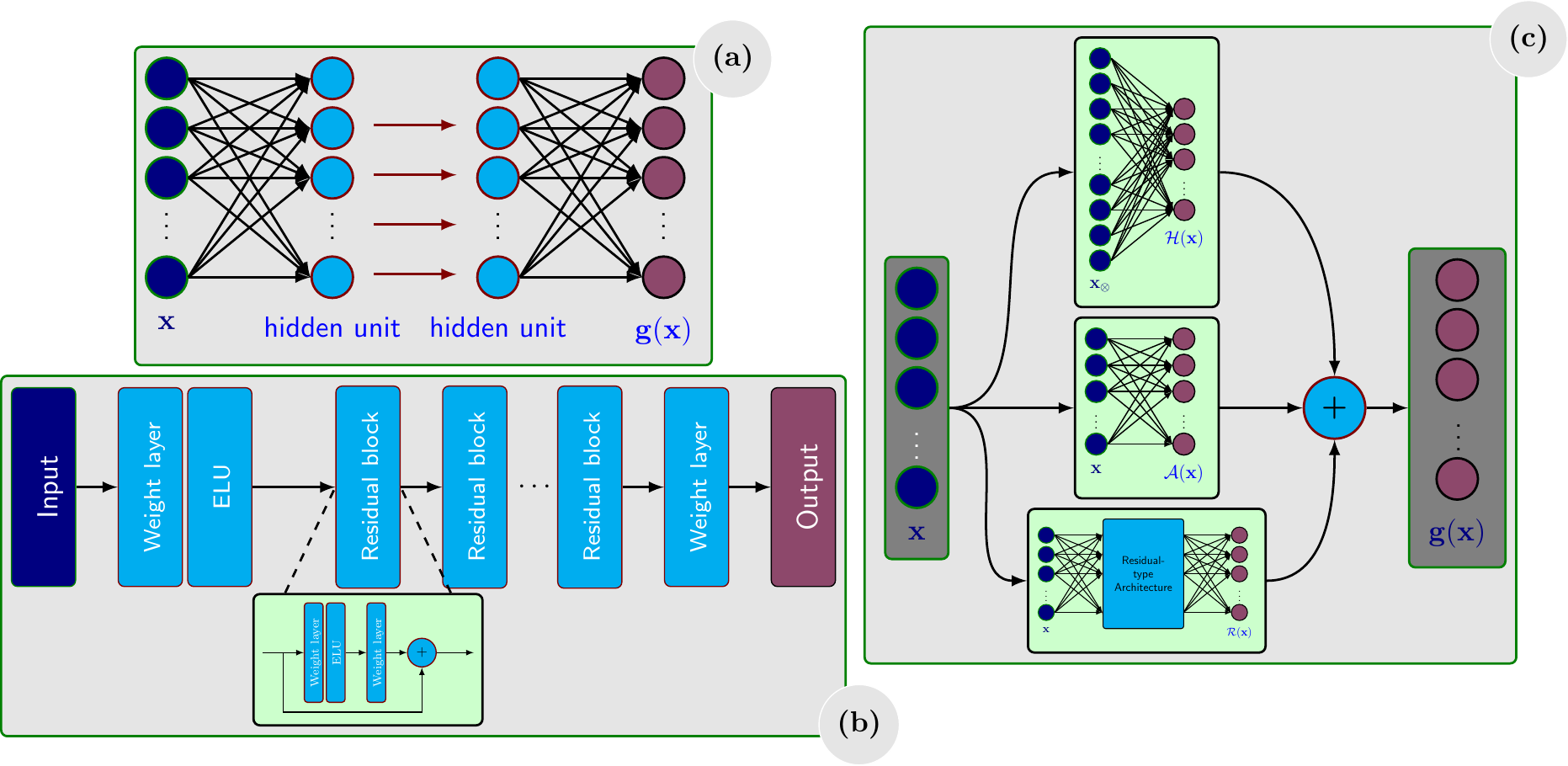}
	\caption{The figure explains different architecture designs of neural networks that essentially aim at mapping the input $\bx$ to $\bg(\bx) (=\dot \bx)$. Sub-figure (a) shows a fully connected deep neural network with $l$ hidden layers; sub-figure (b) illustrates a residual neural network with skip connections that also aims at mapping $\bx$ to $\bg (\bx)$; sub-figure (c) describes the proposed new architecture that, in addition to a residual neural network, has linear and quadratic mappings via skip connections.}
	\label{fig:NN}
\end{figure*}

In this work, we focus on identifying a dynamical model by using not only collected data in simulations or experiments but also available expert knowledge and physical laws about the process. 
In general, we aim at inferring a nonlinear dynamical system of the form:
\begin{equation}\label{eq:NLeqn}
	\dot \bx(t) = \bg(\bx(t)),
\end{equation}
where the vector $\bx(t) \in \Rn$ denotes the state at time $t$, and the function $\bg(\cdot): \Rn \rightarrow \Rn$ is a continuous differentiable function, describing the dynamics of the system. We shall later on discuss its extension to dynamical systems with control and parameters, and its discrete version. 
In principle, we are interested in determining the function $\bg$ from data. To that aim, we observe time-evaluation of the state vector $\bx(t)$ and the derivative $\dot \bx(t)$ (if possible); otherwise, we approximate the derivative information using the state vector $\bx(t)$ by employing, for example, a five-point stencil method. 
DL approaches have been used to learn dynamic models for decades \cite{gonzalez1998identification, milano2002neural}. Precisely, they are used to learn the mapping $\bx(t) \rightarrow \dot \bx(t)$. To that aim, typically, fully connected DNNs can be utilized as shown in Fig.~\ref{fig:NN}(a); however, they share a common drawback --  that is, they are hard to train as the network goes deeper. This is primarily due to gradient vanishing issues that commonly occur in such networks. Consequently, there are little or no updates in the network parameters, yet the network being far from representing the input-output mapping. 
The gradient vanishing problem can be solved to some extent by a \resnet-type architecture. Fundamentally, in \resnets, the output of a layer is passed not only to the next layer but also directly to deeper layers by using skip connections. This is illustrated in Fig.~\ref{fig:NN}(b).  
Furthermore, in the direction of scientific machine learning, it is also important to make use of the additional knowledge which may be available from experts or underlying physical laws. Towards this, operator inference (\opinf) approaches have emerged as potential ones \cite{peherstorfer2016data,morBenGKPW20,morBenGHetal20}. A key observation is that the rate of change of the dependent variable $\bx$ strongly depends linearly or quadratically on the state $\bx$. This observation is often encountered in engineering and biology problems, e.g., flow problems, Fisher's equation for gene propagation \cite{fisher1937wave}, Lorenz system, or FitzHugh-Nagumo \cite{fitzhugh1955mathematical} describing neural dynamics. It has also been shown in \cite{QKPW2020_lift_and_learn} that for sufficiently smooth nonlinear systems, it is possible to define hand-engineering features (called lifted variables) such that the dynamics of a process in the lifted variables can approximately be given in the quadratic form of $\bg(\bx(t))$. 
Moreover, recently, the authors in \cite{mcquarrie2020data} have shown that considering only linear and quadratic dependencies between variables, the dynamics of a single-injector combustion process can be described quite accurately. In this case, one learns dynamical systems of the form:
\begin{equation}\label{eq:DS_quad}
	\dot \bx(t)  =  \bA \bx(t)  + \bQ \left(\bx(t)\otimes \bx(t)\right) + \bb,
\end{equation}
which is typically considered in \opinf~\cite{peherstorfer2016data}, and  $`\otimes'$ denotes the Kronecker product, i.e., $$\bx(t)\otimes \bx(t) = \begin{bmatrix} \bx_1^2(t), \bx_1(t)\bx_2(t), \ldots,\bx_1(t)\bx_n(t),\ldots, \bx_n^2(t) \end{bmatrix},$$ 
where $\bx_i(t)$ denotes the $i$th component of the vector $\bx(t)$.  Furthermore, the system in Eq.~\ref{eq:DS_quad} can also be obtained by considering only terms up-to second order of the Tailor series expansion of $\bg(\bx(t))$ in Eq.~\ref{eq:NLeqn}, i.e., 
\begin{equation}
	\bg(\bx(t))  = \bA\bx(t)  + \bQ \left(\bx(t)\otimes \bx(t)\right) + \bb + \cO(\bx(t)^3).
\end{equation}
If $\bx(t)$ is small, then the higher-order terms are very small, thus can be neglected. However, if this is not the case, then the higher-order terms would still contribute significantly. Hence, in this paper, we consider a model of a dynamical system of the form:
\begin{equation}\label{eq:DS}
	\dot \bx(t)  =  \bA \bx(t)  + \bQ \left(\bx(t)\otimes \bx(t)\right) + \boldf(\bx), \quad \bx(0) = \bx_0.
\end{equation}
To learn a model, having the form as in Eq.~\ref{eq:DS}, we propose the architecture shown in Fig.~\ref{fig:NN}(c). The proposed architecture resembles a \resnet, where linear and quadratic connections are directly passed to the output layer, and  $\boldf(\bx)$ is also a \resnet~in itself. We refer to the architecture as \emph{Linear-Quadratic-Residual Network} (\lqresnet).
Naturally, we can feed directly $\bx$, containing all involved variables, to learn a function $\bg(\bx) :=  \bA \bx(t)  + \bQ \left(\bx(t)\otimes \bx(t)\right) + \boldf(\bx)$ in the proposed architecture; however, we empirically make important observations -- these are:
\begin{itemize}
	\item For $\bx_i(t)$ representing different quantities (e.g., concentrations of species in biology/chemical processes),  it is efficient to build a network for each component, precisely the mapping $\bx(t) \rightarrow \dot \bx_i(t)$, instead of a bigger network, learning the mapping $\bx(t) \rightarrow \dot \bx(t)$. A reason behind this is that a deep network aims at learning the features describing the mapping. Typically, we may require very different features to learn the mapping $\bx(t) \rightarrow \dot \bx_i(t)$, $i\in\{1,\ldots,n\}$. On the other hand, if we learn the mapping $\bx(t) \rightarrow \dot \bx(t)$  at once, then the network needs to learn many complex features at once, which can require a larger network, and we may require a big training data set and longer training time.
	\item Moreover, the mappings $\bx(t) \rightarrow \dot \bx_i(t)$  can be of different complexities; it means that the number of features that need to be learned by a network to describe the mapping can be different. As a result, some of the networks can be smaller/larger than others. 
	\item Moreover, if several small networks are built for each mapping, then we can easily parallelize the forward propagation to predict the output $\dot \bx(t)$ for a given input $\bx(t)$. 
\end{itemize}
Thus, we train a network, describing  the mapping $\bx(t) \rightarrow \dot \bx_i(t)$ by incorporating the linear and quadratic terms. Hence, we aim at solving the following optimization problem:
\begin{equation}\label{eq:optproblem}
	\min_{\bA_j,\bQ_j,\Theta_i} \sum_{j = 1}^\cN \left\|\dot \bx^{(j)}_i - \bA_j \bx^{(j)} - \bQ_j \left(\bx^{(j)} \otimes \bx^{(j)}\right) - \mathcal R_{\Theta_i}(\bx^{(j)})\right\|,
\end{equation}
where  $\bx^{(j)}$ is a state vector at a time instance, and $\dot\bx^{(j)}_i$ denotes the time-derivative of $\bx^{(j)}_i$ at $\bx^{(j)}$.  $\cN$ is the total number of training data, and $\cR_{\Theta_i}$ is a \resnet~(learning $\boldf_i(\bx)$ in Eq.~\ref{eq:DS} and $\boldf_i$ denoting its $i$th entry), parameterized by $\Theta_i$.  

Note that the optimization problem in Eq.\ \ref{eq:optproblem} is high-dimensional and non-convex; hence, it is a hard problem to solve in theory. Moreover, it may have many local minima. However, as illustrated in \cite{visualloss}, neural networks with skip connections such as the proposed one tend to find a global minimum (or get close to it) even with simple stochastic gradient methods.  Moreover, such a network does not suffer from the gradient vanishing problem as well in training. 

Furthermore, in the following, we state our empirical finding.  To solve Eq.~\ref{eq:optproblem}, one may also think of decoupling the optimization problem as follows. In the first step, one can aim at identifying the linear and quadratic matrices by solving the following optimization problem: 
\begin{equation}
	\min_{\bA_j,\bQ_j} \sum_{j=1}^\cN\left\|\dot \bx^{(j)}_i - \bA_j \bx^{(j)} - \bQ_j \left(\bx^{(j)} \otimes \bx^{(j)}\right)\right\|,
\end{equation}
Once we obtain $\bA_i$ and $\bQ_i$, we can define the residual as follows:
\begin{equation}\label{eq:residual}
	\br^{(j)}_i(\bx) = \dot \bx^{(j)}_i - \bA_i \bx^{(j)} - \bQ_i \left(\bx^{(j)}\otimes \bx^{(j)}\right),
\end{equation}
which can then be learned using a \resnet. But in our numerical experimental studies, we observe that this yields a poor model, and it seems to be harder for the network to learn the underlying dynamics. 

Therefore, we suggest to optimize simultaneously the matrices $\bA_j,\bQ_j$, and the parameters for the \resnet~$\cR_{\Theta_i}$ that solves Eq.~\ref{eq:optproblem}. 
However, the quantity $\br^{(j)}_i(\bx)$ can indicate how well the dynamics can be captured only by linear and quadratic terms. Hence, if  $\br^{(j)}_i(\bx)$ is smaller than a threshold, then we do not need to train a network, and $\bA_j$ and $\bQ_i$ can be given analytically; hence, we can obtain a parsimonious interpretable model in this case. 
\subsection*{Interpretation of the LQResNet}
In what follows, we discuss an interpretation of the \lqresnet. For this, we first note that the residual block can be seen as composite functions, i.e., \[\bW\left(\bh_l(\bh_{l-1} \cdots \bh_2(\bh_1(\bx)))\right) + \bb,\] where $\bh_i(\bx) := \bx + \bW^{(2)}_i\mathrm\sigma(\bW^{(1)}_i\bx +\bb^{(1)}) +\bb^{(2)}$, and $\bW,\bW_i^{(1)}, \bW_i^{(2)}$ are weights, $\bb,\bb_i^{(1)},\bb_i^{(2)}$ are the biases, and $\mathrm \sigma(\cdot)$ is a nonlinear activation function.  If we closely look at the network, then we notice that we essentially learn the mapping $\bx(t) \rightarrow \dot \bx_i(t)$, and the mapping takes the form:
\begin{equation}
	\bA_i\left(\bx + \Xi(\bx)\right) + \bQ_i(\bx\otimes \bx).
\end{equation}
Hence, we can think of learning a perturbation to the linear mapping such that it can describe the mapping $\bx(t) \rightarrow \dot \bx_i(t)$. If $\Xi(\bx)$ is only a small perturbation, then it can be learned using a small network; otherwise, we need a bigger one. 
Additionally, the residual blocks in \resnets~to learn $ \Xi(\bx)$ can be adaptively be enlarged in the course of the  training process if the mapping is not sufficiently accurate. This is possible due to the residual blocks in the network that increasingly learns complex features. Therefore, we can expect a parsimonious compact network to get the desired mapping. 
%

In many scenarios, the dynamics of a physical system are governed by partial differential equations. As a result, if data are collected using a black-box simulator or in an experimental set-up on a spatial grid, then the state vector $\bx$ lies in high-dimensional space. For example, consider the Cahn-Hilliard equation that describes the dynamics of the phase separation in binary alloys. To capture the dynamics accurately even in a 2-dimensional space, several thousands of variables may be required. As the state vector dimension increases, the optimization problem (Eq.~\ref{eq:optproblem}) becomes computationally expensive as the number of parameters in the optimization problem is of $\cO(n^2)$, where $n$ is the dimension of the state vector.  However, to ease the problem, we can use the fact that many high-dimensional dynamical systems evolve in a low-dimensional manifold. Thus, the state vector $\bx$ can be well-approximated using $\hn$ basis functions which can be determined using dimensional reduction techniques. A  widely used technique to determine such a basis is proper orthogonal decomposition \cite{kunisch2002galerkin, berkooz1993proper}. In principle, the high-dimensional state vector $\bx(t)$ can be approximated as
\begin{equation}
	\bx(t) \approx \bV \hat \bx(t),
\end{equation}
where $\bV \in \R^{n\times \hn}$, $\bx(t) \in \R^{\hn}$, and $\bV$ is constructed using left singular vectors of the collected measurements of $\bx(t)$, corresponding to ${\hn}$ leading singular values. As a results, we can rather learn a dynamical system for the low-dimensional variable $\hat\bx(t)$:
\begin{equation}
	\dot{\hat\bx}(t) = \hat\bA\hat\bx(t) + \hat\bQ\left(\hat\bx(t)\otimes \hat\bx(t)\right) +  \hat\boldf(\hat\bx(t)).
\end{equation}
For this, we would require data $\hat\bx(t_i)$ and $\dot{\hat\bx}(t_i)$ which can be obtained by projecting the high-dimensional $\bx(t_i)$ and $\dot{\bx}(t_i)$ using the projection matrix $\bV$, i.e., $\hat\bx(t_i) = \bV^\top \bx(t_i)$ and $\dot{\hat\bx}(t_i) = \bV^\top \dot{\bx}(t_i)$.
Furthermore, when $\bx(t)$ denotes variables obtained from a discretization of a partial differential equation, all these variables are alike, meaning they are of similar complexity. Hence, we can directly learn the mapping $\hat \bx(t) \rightarrow \dot{\hat\bx}(t)$ using a single network with linear and quadratic skip connections to the last layer (as depicted in Fig.~\ref{fig:NN}(c)).  Moreover, if the data represents different types of variables -- it happens when dynamics is governed by a couple of partial differential equations (e.g., shallow-water equations \cite{bresch2009shallow}) -- then we train a network to learn the mapping for each class of variables. 

\subsection*{Possible extensions}
Here, we briefly discuss some possible extensions of the above methodology to learn dynamical models. 
\subsubsection*{Additional prior knowledge} There are various engineering applications where we have more insight into the process that may be known by experts or from first principles, for example, an interconnected topological network in biological species, chemical kinetics reactions, flow problems, and mechanical instruments. 
We illustrate this with two examples. In the first example, consider the data describing the motion of an inverted pendulum is given. In this case, the first-principle  information of a motion of an inverted pendulum can be used -- these are, e.g., that the dynamics are of second-order in time and typically contain a sine function, i.e.,
\begin{equation}
	\ddot \bx(t) = \ba\dot \bx(t) +  \bb\sin(\bx(t)),
\end{equation} 
where $\ba,\bb$ are constants. However, in practice, there is some additional mechanical friction that is hard to describe analytically. Thus, one may aim at learning the additional friction factor using a residual network and directly add a sine connection to the output layer to make the learning more efficient. Precisely, we can define an optimization problem as follows:
\begin{equation}
	\min_{\ba,\bb,\Theta}\sum_{j=1}^{\cN}	\|	\ddot \bx^{(j)} - \ba\dot \bx^{(j)} -  \bb\sin(\bx^{(j)}) - \cR_\Theta\left(\dot \bx^{(j)}, \bx^{(j)}\right)\|,
\end{equation}
where $\cR_\Theta$ is a \resnet, parameterized by $\Theta$. Having set up this, we can expect a far better model with a smaller amount of data for an inverted pendulum that describes the dynamics as close to reality as possible.  In the second example, let us assume the data represents incompressible flow dynamics and denote the velocity field $\bv(t)$ at time $t$. If we aim at learning a model describing the dynamics of the flow, we need to ensure that the model enforces the incompressibility condition. For this, we project the velocity field $\bv(t)$ on a manifold, where the condition is met. We can find such a manifold using the singular value decomposition of the data matrix containing the velocity field, i.e., $\hat\bv(t) = \bV^\top\bv(t)$. Then, we build an \lqresnet, describing the dynamics of $\hat\bv(t)$ and the velocity field $\bv(t) = \bV\hat\bv(t)$ which will inherently satisfy the mass-conservation law.

\subsubsection*{Discrete models}
One may also be interested in learning a discrete model which takes the form:
\begin{equation}
	\bx(k+1) =  \bA\bx(k) + \bQ(\bx(k)\otimes \bx(k)) + \bG(\bx(k)),
\end{equation}
where $\bx(k) \in \Rn$ is the state $\bx$ at the $k$th time step;  $\bA \in \Rnn, \bQ \in \R^{n\times n^2}$, and the nonlinear function $\bG(\cdot): \Rn \rightarrow \Rn$.   In this case, we can readily apply the approach discussed above that takes the state  $\bx$ at  time step $k$ as an input and yields  the state at the next time step as an output. An advantage of the approach is that we do not require the derivative information of the state $\bx$. The derivative information can be challenging to estimate when the noise in the state is above a threshold level with classical difference methods. 
However, there exist approaches to estimate derivative information using, e.g., total variation regularization \cite{chartrand2011numerical}, or one can employ the deep learning-based technique \cite{rudy2019deep} can be used to remove the first noise from a signal and then apply finite-difference or polynomial based method to estimate derivative. 

\subsubsection*{Controlled and parametric case}
Assume that the dynamic process is subject to an external input and involves parameter involved. In this case, inspired by our earlier discussion, we aim at learning a dynamical system of the form:
\begin{equation}
	\begin{aligned}
		\dot{\bx}(t,\mu) &= \left(\bA + \mu \bA_\mu\right)\bx(t,\mu) + \left(\bQ + \mu \bQ_\mu\right)\left(\bx(t,\mu)\otimes \bx(t,\mu)\right) \\
		&\quad +  \bB\bu(t) + \bR(\bx(t,\mu),\bu(t)),
	\end{aligned}
\end{equation}
where $\mu \in \R, \bu(t) \in \Rm$ denote parameters and control input, respectively; $\bx(t,\mu) \in \Rn$ is the parameter dependent state, and $\bR(\bx(t,\mu),\bu(t)): \R^{n+p}\rightarrow \Rn$. In this case, we can adapt a similar architecture (shown in Fig.~\ref{fig:NN}). Precisely, the adaption is as follows. We have a direct input connection to the output layer, and a residual network can have $\mu$ and $\bu(t)$ along with $\bx(t,\mu)$ as inputs and the whole network is training simultaneously to predict $\dot \bx(t,\mu)$. 

	\section*{Demonstration of the Approach}
We demonstrate the efficiency of the proposed approach to learn dynamical systems using two examples, namely the FitzHugh-Nagumo model and the Glycolytic oscillator. 
We collect data by simulating the governing equations using the Python routine \texttt{odeint} from \texttt{scipy.integrate}. 
Moreover, given data $\bx(t)$ at time instance $\{t_0,\ldots,t_n\}$, we approximate $\dot{\bx}(t)$ using a five-point stencil.
We have used an exponential linear unit as an activation function \cite{clevert2015fast} that does not suffer from dying neuron problem as well as the activation function is $\mathrm C^1$ continuous. 
The weights of a neural network are optimized using a variant of the Adam method \cite{kingma2014adam}, namely rectified Adam \cite{liu2019radam} that utilizes a warm-up strategy for better convergence. Also, to avoid over-fitting, we regularize the optimization by the 2-norm of the weights involved in the network (often referred to as weight decay).
We split the data set into the training and validation data sets in $80{:}20$ ratio. 
Other important hyper-parameters used to train a network for both problems are listed in \Cref{tab:parameters}.
All experiments were performed using PyTorch in Python running on a Macbook Pro
with 2,3 GHz 8-Core Intel Core i9  CPU, 16GB of RAM, and Mac OS X v10.15.6. 

\begin{table}[tb]
	\centering
	\begin{tabular}{|c|c|c|c|c|}\hline
		Problem & Epochs & Learning rate    & Batch size & Weight decay \\ \hline
		FHN     & 500    & $5\cdot 10^{-4}$ & 512        & $10^{-4}$    \\ \hline
		GO      & 2~000   & $10^{-3}$        & 512        & $10^{-4}$   \\ \hline
	\end{tabular}
	\caption{The parameters used to train a neural network are listed. }
	\label{tab:parameters}
\end{table}

\subsection*{FitzHugh-Nagumo Model}
\begin{figure*}
	\includegraphics[width=\textwidth]{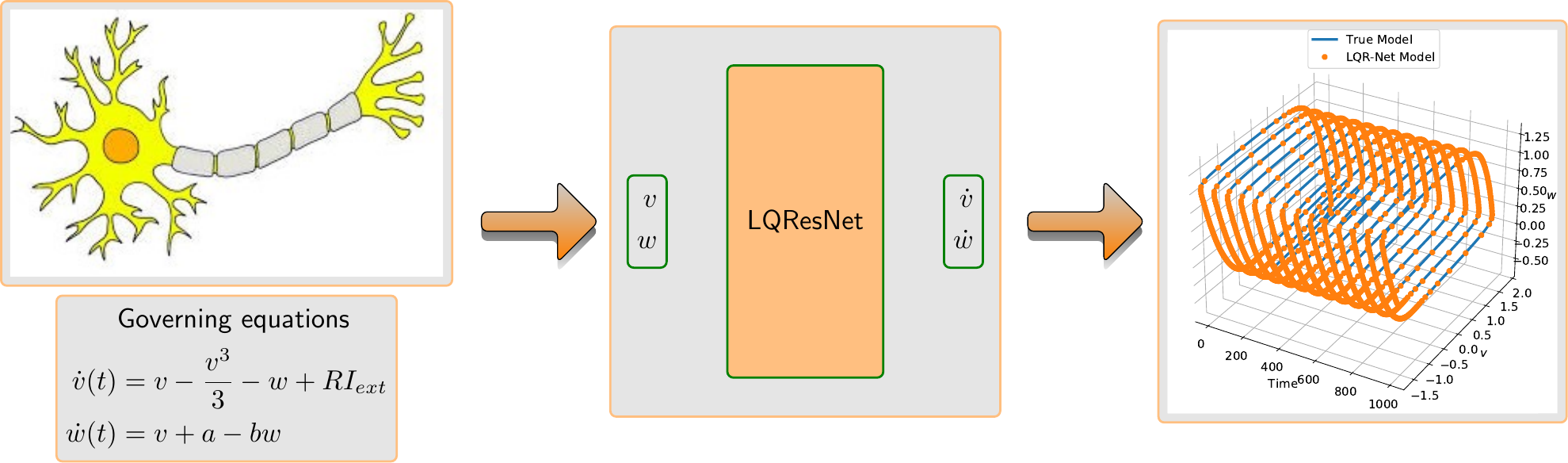}
	\caption{FitzHugh-Nagumo Model: Governing equations, learning model, and comparison of the learned and true models. The left-most figure describing neurons is due to courtesy of~\cite{zhen2019influence}.}
	\label{fig:FHN}
\end{figure*}

As a first example, we consider the FitzHugh-Nagumo model \cite{fitzhugh1961impulses} that describes spiking of a neuron, neuronal dynamical in a simplistic way. The governing equations of such a dynamical behavior are shown in Fig.~\ref{fig:FHN}. The variables $v(t)$ and $w(t)$ describe activation and de-activation of a neuron. Specifically, the model exhibits a periodic oscillatory behavior when $\bI_\texttt{ext}$ exceeds a threshold value. 
We collect data by simulating the model using $10$ randomly chosen initial conditions in the range $[-1,1]\times [-1,1]$, and for each initial condition, we collect  $5~000$ equidistant points in the time interval $[0,200]$s.  
Having these data, we build two models that map $\{v(t),w(t)\} \mapsto \dot{v}(t)$ and $\{v(t),w(t)\} \mapsto \dot w(t)$. 
Before building a deep neural network, we determine the residual as shown in Eq.~\ref{eq:residual}. As a result, we find that $\dot{w}(t)$ can accurately be written in the linear and quadratic forms of $\{v(t),w(t)\}$; thus, we have analytical model for the mapping. On the other hand, the linear and quadratic forms of $\{v(t),w(t)\}$ are not sufficient to represent $\dot v(t)$. Hence, we make use of the residual type neural network to learn the mapping $\{v(t),w(t)\} \mapsto \dot{v}(t)$ using two residual blocks and each hidden layer having ten neurons. Having trained the model, we test the learned model with the true model for an initial condition -- it is neither a part of the training  nor the validation set -- to examine the predictive capacities  of the model. We observe that the learned model replicates the dynamics of the true model very well, even for the unseen initial conditions (see Fig.~\ref{fig:FHN}).

\subsection*{Glycolytic Oscilator}
Our next example represents biochemical dynamics; precisely, it describes the dynamical behavior in yeast glycolysis, see, e.g., \cite{daniels2015efficient}. 
The dynamics are given by the following governing equations that predict the concentrations of 7 different biochemical spices:
\begin{subequations}\allowdisplaybreaks\label{eq:gly_osc}
	\begin{align}
		\dot \bS_1 &= \bJ_0 -\dfrac{k_1\bS_1\bS_6}{1+(\bS_6/k_1)^q},\\
		\dot \bS_2 &= 2\dfrac{k_1\bS_1\bS_6}{1+(\bS_6/k_1)^q} - k_2\bS_2(\bN-\bS_5) - k_6\bS_2\bS_5,\\
		\dot \bS_3 &= k_2\bS_2(\bN-\bS_5) - k_3\bS_3(\bA-\bS_6),\\
		\dot \bS_4 &= k_3\bS_3(\bA-\bS_6) - k_4\bS_4\bS_5 - \kappa(\bS_4-\bS_7),\\
		\dot \bS_5 &= k_2\bS_2(\bN-\bS_5) -k_4\bS_4\bS_5 - k_6\bS_2\bS_5,\\
		\dot \bS_6&= -2\dfrac{k_1\bS_1\bS_6}{1+(\bS_6/k_1)^q} + 2k_3\bS_3(A-\bS_6) - k_5\bS_6,\\
		\dot \bS_7 &= \psi\kappa (\bS_4-\bS_7) - \kappa \bS_7.
	\end{align}
\end{subequations}
We consider the same model parameters as given in \cite[Tab. 1]{daniels2015efficient}, and the range of initial conditions for all concentrations are taken the same as given \cite[Tab. 2]{daniels2015efficient}. In order to collect data, we randomly take $30$ different initial conditions in the given range and for each initial condition, we take  $4~000$ equidistant points in the time interval $[0,10]$. 

\begin{figure*}[tb]
	\def\svgwidth{\textwidth}
\begingroup%
  \makeatletter%
  \providecommand\color[2][]{%
    \errmessage{(Inkscape) Color is used for the text in Inkscape, but the package 'color.sty' is not loaded}%
    \renewcommand\color[2][]{}%
  }%
  \providecommand\transparent[1]{%
    \errmessage{(Inkscape) Transparency is used (non-zero) for the text in Inkscape, but the package 'transparent.sty' is not loaded}%
    \renewcommand\transparent[1]{}%
  }%
  \providecommand\rotatebox[2]{#2}%
  \newcommand*\fsize{\dimexpr\f@size pt\relax}%
  \newcommand*\lineheight[1]{\fontsize{\fsize}{#1\fsize}\selectfont}%
  \ifx\svgwidth\undefined%
    \setlength{\unitlength}{894.65681208bp}%
    \ifx\svgscale\undefined%
      \relax%
    \else%
      \setlength{\unitlength}{\unitlength * \real{\svgscale}}%
    \fi%
  \else%
    \setlength{\unitlength}{\svgwidth}%
  \fi%
  \global\let\svgwidth\undefined%
  \global\let\svgscale\undefined%
  \makeatother%
  \begin{picture}(1,0.52310562)%
    \lineheight{1}%
    \setlength\tabcolsep{0pt}%
    \put(0,0){\includegraphics[width=\unitlength,page=1]{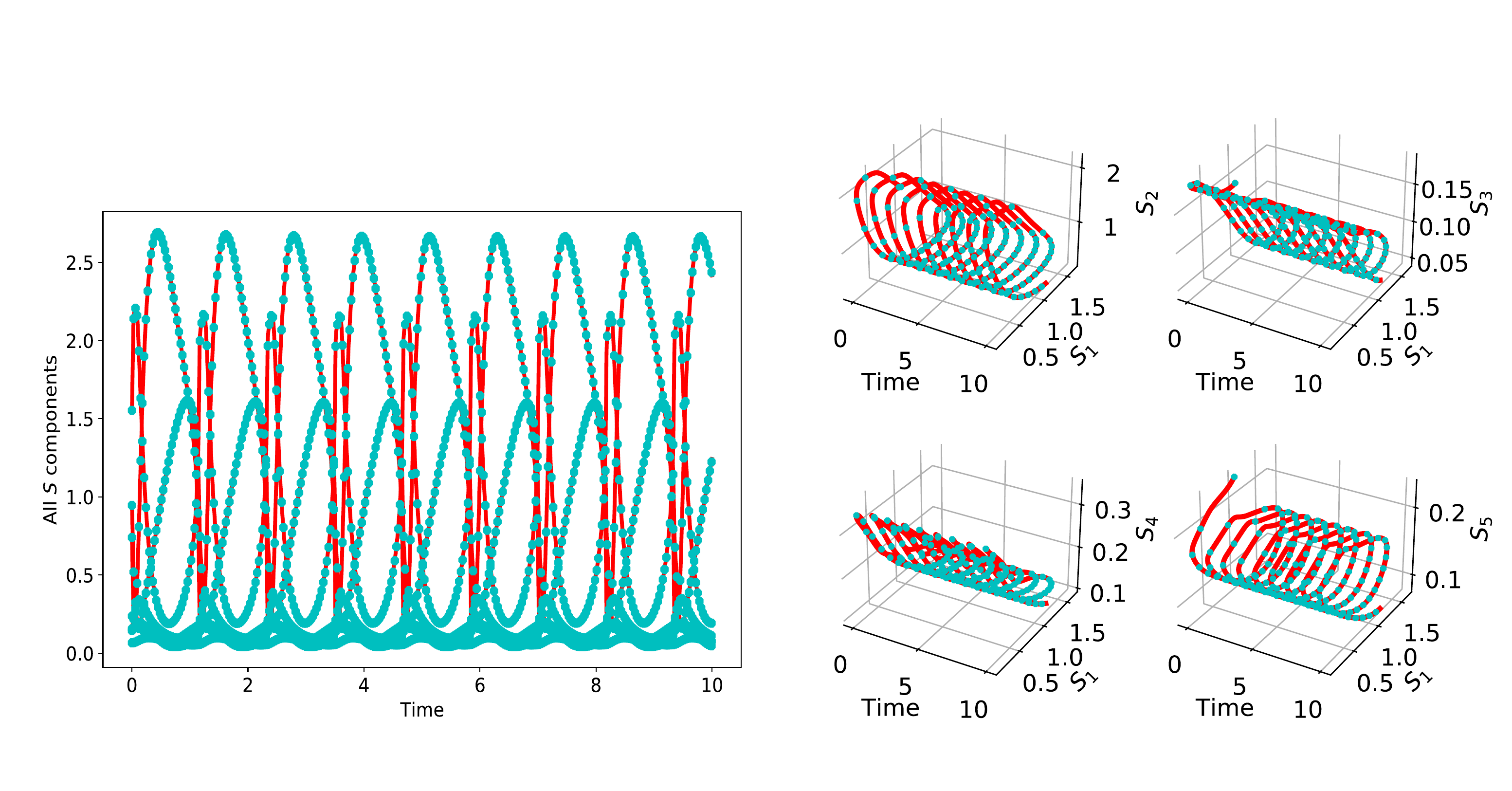}}%
    \put(0.1807329,0.42011316){\color[rgb]{0,0,0}\makebox(0,0)[lt]{\lineheight{1.25}\smash{\begin{tabular}[t]{l}Dynamics of Glycolytic Oscillator \end{tabular}}}}%
        \put(0.0807329,0.12011316){\color[rgb]{0,0,0}\makebox(0,0)[lt]{\lineheight{1.25}\smash{\begin{tabular}[t]{l}$x = y$\end{tabular}}}}%
  \end{picture}%
\endgroup%

	\caption{Glycolytic Oscillator: A comparison of true and learned model for an initial condition that is not used for training. In red color (true model), and cyan dotted (learned model).}
	\label{fig:GO}
\end{figure*}

Clearly, the model shows complicated nonlinear dynamics; thus, it often cannot be known or guessed if only data are provided. 
Albeit complex dynamics, one may notice that the rates of change of concentrations heavily depend on linear and quadratic forms of the concentrations at a given time.
Precisely, linear and quadratic forms of the concentrations at time $t$ with appropriate coefficients can accurately express   $\{\dot \bS_3(t),\dot \bS_4(t),\dot \bS_5(t),\dot \bS_7(t)\}$. Since $\{\dot \bS_1,\dot \bS_2, \dot \bS_6\}$ cannot be accurately given in the linear-quadratic form of the concentrations, we build a   \lqresnet~type deep neural network to learn the mapping $\{\bS_1,\ldots,\bS_7\}\mapsto \dot \bS_i$, $i\in\{1,2,6\}$. For this, the network consists of 5 residual blocks and each hidden layer has $35$ neurons. 
Having built a model, we test the accuracy of the learned model with the true models for an initial condition that has not been considered in the training or validation phase. It is compared in Fig.~\ref{fig:GO}, indicating that the learned model can successfully replicate the dynamics without any prior knowledge of the biochemical process or model. Moreover, we note that the learning process can be even more improved if a topological network describing the interconnection of species is known. To make it clearer, from Eq.~\ref{eq:gly_osc}, we have that the dynamics of $\bS_1$ only  dependent on $\bS_1,\bS_6$; hence, we need to feed only theses variables into a network to learn the dynamics of $\bS_1$. 

\section*{Discussion}
In essence, we have discussed a compelling approach to learning complex nonlinear dynamical systems using data that can incorporate any prior knowledge of a process. The approach, in particular, makes use of the observation, which is often found in dynamical systems' modeling -- that is, the rate of change of a variable strongly depends on linear and quadratic forms of the variable. Thus, we have proposed an efficient deep learning architecture, \lqresnet, that can be seen as a good compromise between purely interpretable mechanistic approaches and black-box neural network approaches. The approach can be employed in various fields, where abundant data can be obtained, where underlying physics or models describing dynamics are not completely known, for instance, in biological process modeling, climate science,  epidemiology, and financial markets.
Moreover, there are various instances where we have a fairly good understanding of the process, which may be derived from physical laws or expert knowledge, but it fails to explain the data collection, for instance, in an experimental setup, due to some hidden dynamics or forces. Hence, the proposed methodology can be applied to learn a correction term using \lqresnet, by combining the knowledge of the process as well as data. As a result, we expect to obtain a good model with a limited amount of data.
We have demonstrated the efficiency of the proposed methodology to learning models using two examples arising in biology and biochemistry. We have observed that these models generalize even to unseen conditions.

One of the current limitations of the approach is the need for accurate data that sufficiently approximates the derivative information. The derivative information is necessary to obtain a continuous-time dynamic model; however, this is challenging to attain from noisy measurements. Indeed, we may first obtain a discrete-time model and then determine an approximate continuous-time model. However, as our future work, we intend to propose efficient deep learning approaches to obtain continuous-time models from noisy data by filtering the noise in the course of learning, though there are some advancements in the direction in \cite{rudy2019deep}. In the future, we also aim at tailoring the approach to the case when data is not collected at a regular time interval. 

\section*{Code Availability}
Our Python implementation using PyTorch is available online under the link
\begin{center} 
	\href{https://github.com/mpimd-csc/LQRes-Net}{https://github.com/mpimd-csc/LQRes-Net}.
	\end{center}


%


\addcontentsline{toc}{section}{References}
\bibliographystyle{plainurl}
\bibliography{mybib}

\begin{thebibliography}{10}

\bibitem{morBenGHetal20}
Peter Benner, Pawan Goyal, Jan Heiland, and Igor Pontes~Duff.
\newblock Operator inference and physics-informed learning of low-dimensional
  models for incompressible flows.
\newblock {arXiv:2010.06701}, 2020.

\bibitem{morBenGKPW20}
Peter Benner, Pawan Goyal, Boris Kramer, Benjamin Peherstorfer, and Karen
  Willcox.
\newblock Operator inference for non-intrusive model reduction of systems with
  non-polynomial nonlinear terms.
\newblock {\em Comp. Meth. Appl. Mech. Eng.}, 372:113433, 2020.

\bibitem{berkooz1993proper}
Gal Berkooz, Philip Holmes, and John~L Lumley.
\newblock The proper orthogonal decomposition in the analysis of turbulent
  flows.
\newblock {\em Annual Rev. Fluid Mech.}, 25(1):539--575, 1993.

\bibitem{bongard2007automated}
Josh Bongard and Hod Lipson.
\newblock Automated reverse engineering of nonlinear dynamical systems.
\newblock {\em Proc. Nat. Acad. Sci. U.S.A.}, 104(24):9943--9948, 2007.

\bibitem{bresch2009shallow}
Didier Bresch.
\newblock Shallow-water equations and related topics.
\newblock In {\em Handbook of Differential Equations: Evolutionary Equations},
  volume~5, pages 1--104. Elsevier, 2009.

\bibitem{brunton2016discovering}
Steven~L Brunton, Joshua~L Proctor, and J~Nathan Kutz.
\newblock Discovering governing equations from data by sparse identification of
  nonlinear dynamical systems.
\newblock {\em Proc. Nat. Acad. Sci. U.S.A.}, 113(15):3932--3937, 2016.

\bibitem{candes2006robust}
Emmanuel~J Cand{\`e}s, Justin Romberg, and Terence Tao.
\newblock Robust uncertainty principles: {E}xact signal reconstruction from
  highly incomplete frequency information.
\newblock {\em IEEE Trans. Inform. Theory}, 52(2):489--509, 2006.

\bibitem{chartrand2011numerical}
Rick Chartrand.
\newblock Numerical differentiation of noisy, nonsmooth data.
\newblock {\em ISRN Appl. Math.}, 2011, 2011.

\bibitem{chen2018neural}
Ricky~TQ Chen, Yulia Rubanova, Jesse Bettencourt, and David~K Duvenaud.
\newblock Neural ordinary differential equations.
\newblock In {\em Advances Neural Inform. Processing Sys.}, pages 6571--6583,
  2018.

\bibitem{clevert2015fast}
Djork-Arn{\'e} Clevert, Thomas Unterthiner, and Sepp Hochreiter.
\newblock Fast and accurate deep network learning by exponential linear units
  {(ELUs)}.
\newblock {\em arXiv preprint arXiv:1511.07289}, 2015.

\bibitem{crutchfield1987equations}
James~P Crutchfield and Bruce~S McNamara.
\newblock Equations of motion from a data series.
\newblock {\em Complex Sys.}, 1(417-452):121, 1987.

\bibitem{daniels2015automated}
Bryan~C Daniels and Ilya Nemenman.
\newblock Automated adaptive inference of phenomenological dynamical models.
\newblock {\em Nature Comm.}, 6(1):1--8, 2015.

\bibitem{daniels2015efficient}
Bryan~C Daniels and Ilya Nemenman.
\newblock Efficient inference of parsimonious phenomenological models of
  cellular dynamics using {S}-systems and alternating regression.
\newblock {\em {PLoS One}}, 10(3):e0119821, 2015.

\bibitem{donoho2006compressed}
David~L Donoho.
\newblock Compressed sensing.
\newblock {\em IEEE Trans. Inform. Theory}, 52(4):1289--1306, 2006.

\bibitem{fisher1937wave}
Ronald~Aylmer Fisher.
\newblock The wave of advance of advantageous genes.
\newblock {\em Annals of Eugenics}, 7(4):355--369, 1937.

\bibitem{fitzhugh1955mathematical}
Richard FitzHugh.
\newblock Mathematical models of threshold phenomena in the nerve membrane.
\newblock {\em The Bulletin Math. Biophys.}, 17(4):257--278, 1955.

\bibitem{fitzhugh1961impulses}
Richard FitzHugh.
\newblock Impulses and physiological states in theoretical models of nerve
  membrane.
\newblock {\em Biophysical J.}, 1(6):445--466, 1961.

\bibitem{gonzalez1998identification}
R~Gonzalez-Garcia, R~Rico-Martinez, and IG~Kevrekidis.
\newblock Identification of distributed parameter systems: {A} neural net based
  approach.
\newblock {\em Computers \& Chemical Engrg.}, 22:S965--S968, 1998.

\bibitem{goodfellow2016deep}
Ian Goodfellow, Yoshua Bengio, and Aaron Courville.
\newblock {\em Deep learning}.
\newblock MIT press, 2016.

\bibitem{he2016deep}
Kaiming He, Xiangyu Zhang, Shaoqing Ren, and Jian Sun.
\newblock Deep residual learning for image recognition.
\newblock In {\em Proc. IEEE Conf. Comp. Vision Patt. Recog.}, pages 770--778,
  2016.

\bibitem{kantz2004nonlinear}
Holger Kantz and Thomas Schreiber.
\newblock {\em Nonlinear Time Series Analysis}, volume~7.
\newblock Cambridge University Press, 2004.

\bibitem{kevrekidis2003equation}
Ioannis~G Kevrekidis, C~William Gear, James~M Hyman, Panagiotis~G Kevrekidis,
  Olof Runborg, Constantinos Theodoropoulos, et~al.
\newblock Equation-free, coarse-grained multiscale computation: {E}nabling
  mocroscopic simulators to perform system-level analysis.
\newblock {\em Comm. Math. Sci.}, 1(4):715--762, 2003.

\bibitem{kingma2014adam}
Diederik~P Kingma and Jimmy Ba.
\newblock Adam: A method for stochastic optimization.
\newblock {\em arXiv preprint arXiv:1412.6980}, 2014.

\bibitem{krizhevsky2012imagenet}
Alex Krizhevsky, Ilya Sutskever, and Geoffrey~E Hinton.
\newblock Imagenet classification with deep convolutional neural networks.
\newblock In {\em Adv. Neural Inform. Processing Sys.}, pages 1097--1105, 2012.

\bibitem{NarP90}
S~Narendra Kumpati and Parthasarathy Kannan.
\newblock Identification and control of dynamical systems using neural
  networks.
\newblock {\em IEEE Trans. Neural Networks}, 1(1):4--27, 1990.

\bibitem{kunisch2002galerkin}
Karl Kunisch and Stefan Volkwein.
\newblock Galerkin proper orthogonal decomposition methods for a general
  equation in fluid dynamics.
\newblock {\em {SIAM} J. Numer. Anal.}, 40(2):492--515, 2002.

\bibitem{lecun2015deep}
Yann LeCun, Yoshua Bengio, and Geoffrey Hinton.
\newblock Deep learning.
\newblock {\em Nature}, 521(7553):436--444, 2015.

\bibitem{visualloss}
Hao Li, Zheng Xu, Gavin Taylor, Christoph Studer, and Tom Goldstein.
\newblock Visualizing the loss landscape of neural nets.
\newblock In {\em Neural Inform. Processing Syst.}, 2018.

\bibitem{liu2019radam}
Liyuan Liu, Haoming Jiang, Pengcheng He, Weizhu Chen, Xiaodong Liu, Jianfeng
  Gao, and Jiawei Han.
\newblock On the variance of the adaptive learning rate and beyond.
\newblock {\em arXiv preprint arXiv:1908.03265}, 2019.

\bibitem{lennart1999system}
Lennart Ljung.
\newblock {\em System Identification: Theory for the User}.
\newblock Prentice Hall, NJ, 1999.

\bibitem{mcquarrie2020data}
Shane~A McQuarrie, Cheng Huang, and Karen Willcox.
\newblock Data-driven reduced-order models via regularized operator inference
  for a single-injector combustion process.
\newblock {\em arXiv preprint arXiv:2008.02862}, 2020.

\bibitem{milano2002neural}
Michele Milano and Petros Koumoutsakos.
\newblock Neural network modeling for near wall turbulent flow.
\newblock {\em J. Comput. Phys.}, 182(1):1--26, 2002.

\bibitem{ozolicnvs2013compressed}
Vidvuds Ozoli{\c{n}}{\v{s}}, Rongjie Lai, Russel Caflisch, and Stanley Osher.
\newblock Compressed modes for variational problems in mathematics and physics.
\newblock {\em Proc. Nat. Acad. Sci. U.S.A.}, 110(46):18368--18373, 2013.

\bibitem{peherstorfer2016data}
Benjamin Peherstorfer and Karen Willcox.
\newblock Data-driven operator inference for nonintrusive projection-based
  model reduction.
\newblock {\em Comp. Meth. Appl. Mech. Eng.}, 306:196--215, 2016.

\bibitem{proctor2014exploiting}
Joshua~L Proctor, Steven~L Brunton, Bingni~W Brunton, and JN~Kutz.
\newblock Exploiting sparsity and equation-free architectures in complex
  systems.
\newblock {\em Europ. Phy. J. Spec. Top.}, 223(13):2665--2684, 2014.

\bibitem{QKPW2020_lift_and_learn}
E.~Qian, B.~Kramer, B.~Peherstorfer, and K.~Willcox.
\newblock Lift \& learn: Physics-informed machine learning for large-scale
  nonlinear dynamical systems.
\newblock {\em Physica {D}: {N}onlinear {P}henomena}, 406:132401, 2020.
\newblock URL: \url{https://doi.org/10.1016/j.physd.2020.132401}.

\bibitem{raissi2019physics}
Maziar Raissi, Paris Perdikaris, and George~E Karniadakis.
\newblock Physics-informed neural networks: A deep learning framework for
  solving forward and inverse problems involving nonlinear partial differential
  equations.
\newblock {\em J. Comput. Phys.}, 378:686--707, 2019.

\bibitem{rowley2009spectral}
Clarence~W Rowley, Igor Mezi\'c, Shervin Bagheri, Philipp Schlatter, and Dans
  Henningson.
\newblock Spectral analysis of nonlinear flows.
\newblock {\em J. Fluild Mech.}, 641(1):115--127, 2009.

\bibitem{rudy2017data}
Samuel~H Rudy, Steven~L Brunton, Joshua~L Proctor, and J~Nathan Kutz.
\newblock Data-driven discovery of partial differential equations.
\newblock {\em Sci. Adv.}, 3(4):e1602614, 2017.

\bibitem{rudy2019deep}
Samuel~H Rudy, J~Nathan Kutz, and Steven~L Brunton.
\newblock Deep learning of dynamics and signal-noise decomposition with
  time-stepping constraints.
\newblock {\em J. Comput. Phys.}, 396:483--506, 2019.

\bibitem{schaeffer2013sparse}
Hayden Schaeffer, Russel Caflisch, Cory~D Hauck, and Stanley Osher.
\newblock Sparse dynamics for partial differential equations.
\newblock {\em Proc. Nat. Acad. Sci. U.S.A.}, 110(17):6634--6639, 2013.

\bibitem{schmid2010dynamic}
Peter~J Schmid.
\newblock Dynamic mode decomposition of numerical and experimental data.
\newblock {\em J. Fluild Mech.}, 656:5--28, 2010.

\bibitem{schmidt2011automated}
Michael~D Schmidt, Ravishankar~R Vallabhajosyula, Jerry~W Jenkins, Jonathan~E
  Hood, Abhishek~S Soni, John~P Wikswo, and Hod Lipson.
\newblock Automated refinement and inference of analytical models for metabolic
  networks.
\newblock {\em Phy. Biology}, 8(5):055011, 2011.

\bibitem{SuyVdM96}
Johan~AK Suykens, Joos~PL Vandewalle, and Bart~L de~Moor.
\newblock {\em Artificial Neural Networks for Modelling and Control of
  Non-Linear Systems}.
\newblock Springer, 1996.

\bibitem{tibshirani1996regression}
Robert Tibshirani.
\newblock Regression shrinkage and selection via the lasso.
\newblock {\em J. Royal Stat. Soc.: Series B (Methodological)}, 58(1):267--288,
  1996.

\bibitem{VanM96}
Peter {Van~Overschee} and Bart {de Moor}.
\newblock {\em Subspace Identification of Linear Systems: Theory,
  Implementation, Applications}.
\newblock Kluwer Academic Publishers, 1996.

\bibitem{wang2011predicting}
Wen-Xu Wang, Rui Yang, Ying-Cheng Lai, Vassilios Kovanis, and Celso Grebogi.
\newblock Predicting catastrophes in nonlinear dynamical systems by compressive
  sensing.
\newblock {\em Phy. Rev. Letters}, 106(15):154101, 2011.

\bibitem{williams2015data}
Matthew~O Williams, Ioannis~G Kevrekidis, and Clarence~W Rowley.
\newblock A data--driven approximation of the {K}oopman operator: Extending
  dynamic mode decomposition.
\newblock {\em J. Nonlinear Sci.}, 25(6):1307--1346, 2015.

\bibitem{ye2015equation}
Hao Ye, Richard~J Beamish, Sarah~M Glaser, Sue~CH Grant, Chih-hao Hsieh,
  Laura~J Richards, Jon~T Schnute, and George Sugihara.
\newblock Equation-free mechanistic ecosystem forecasting using empirical
  dynamic modeling.
\newblock {\em Proc. Nat. Acad. Sci. U.S.A.}, 112(13):E1569--E1576, 2015.

\bibitem{zhen2019influence}
Bin Zhen, Zhenhua Li, and Zigen Song.
\newblock Influence of time delay in signal transmission on synchronization
  between two coupled {F}itz{H}ugh-{N}agumo neurons.
\newblock {\em Appl. Sci.}, 9(10):2159, 2019.

\end{thebibliography}
  
\end{document}